\documentclass[]{interact}

\usepackage{epstopdf}
\usepackage[caption=false]{subfig}

\usepackage[numbers,sort&compress]{natbib}

\usepackage{here}
\usepackage{mathtools}
\usepackage{url}
\usepackage{hyperref}

\bibpunct[, ]{[}{]}{,}{n}{,}{,}
\renewcommand\bibfont{\fontsize{10}{12}\selectfont}
\makeatletter
\def\NAT@def@citea{\def\@citea{\NAT@separator}}
\makeatother

\theoremstyle{plain}

\theoremstyle{definition}

\theoremstyle{remark}

\begin{document}

\title{Viscoelasticity Estimation of Sports Prosthesis by Energy-minimizing Inverse Kinematics and Its Validation by Forward Dynamics}

\author{
\name{Yuta Shimane\textsuperscript{a}, Taiki Ishigaki\textsuperscript{a}, Sunghee Kim\textsuperscript{a} and Ko Yamamoto\textsuperscript{a}}
\affil{\textsuperscript{a}Department of Mechano-informatics, The University of Tokyo, 7-3-1 Hongo, Bunkyo-ku, Tokyo 113-8656, Japan}
}
\thanks{
© 2024 Copyright held by the owner/author(s).
This is an original manuscript of an article published by Taylor \& Francis in Advanced Robotics on 4th Oct. 2024, available at: \url{https://doi.org/10.1080/01691864.2024.2407118}.
}

\maketitle

\begin{abstract}
In this study, we present a method for estimating the viscoelasticity of a leaf-spring sports prosthesis using advanced energy minimizing inverse kinematics based on the Piece-wise Constant Strain (PCS) model to reconstruct the three-dimensional dynamic behavior.
Dynamic motion analysis of the athlete and prosthesis is important to clarify the effect of prosthesis characteristics on foot function. However, three-dimensional deformation calculations of the prosthesis and viscoelasticity have rarely been investigated.
In this letter, we apply the PCS model to a prosthesis deformation, which can calculate flexible deformation with low computational cost and handle kinematics and dynamics. In addition, we propose an inverse kinematics calculation method that is consistent with the material properties of the prosthesis by considering the minimization of elastic energy. Furthermore, we propose a method to estimate the viscoelasticity by solving a quadratic programming based on the measured motion capture data.
The calculated strains are more reasonable than the results obtained by conventional inverse kinematics calculation.
From the result of the viscoelasticity estimation, we simulate the prosthetic motion by forward dynamics calculation and confirm that this result corresponds to the measured motion.
These results indicate that our approach adequately models the dynamic phenomena, including the viscoelasticity of the prosthesis.
\end{abstract}

\begin{keywords}
Prosthesis; Viscoelasticity; Optimization; Kinematics; Dynamics
\end{keywords}


\section{Introduction}
The sports prosthesis plays an important role in enabling people with disabilities to demonstrate their individuality and abilities and improve their performance. 
For example, the 100-metre sprint time at the Paralympics for individuals with a lower extremity amputation who wear carbon-fibre running prostheses has been improved with technology. In the future, they could become the fastest sprinters in the world regardless of disability \cite{hobara2015}.
However, it is not easy to design, as it requires repeated prototyping and trial and error based on experience. It is also unclear how the stiffness of the prosthesis affects muscle and foot function. \\ \indent
Motion analysis is important for understanding how a prosthesis affects the body and provides valuable information for rehabilitation and training. 
In particular, the integration of robotic technology, such as a human skeletal model, has advanced our understanding of biomechanical body function. 
This approach is also effective for investigating the relationship between prosthesis and human, with several studies \cite{price2019, guzelbulut2021, ueyama2020}. 
The introduction of a multi-link model of a prosthesis into a whole-body skeletal model, defined as a rigid multi-link system, suggests the effects of prosthesis geometry and internal parameters on the kinematics and dynamics of joints and energy during walking \cite{rigney2016}. The effects of prosthesis stiffness on muscle and foot function were suggested by simulating the dynamics of an amputee's gait by varying the viscoelasticity of the prosthesis \cite{fey2013}. However, in these studies, inverse dynamics calculations assuming a rigid body do not take into account the elastic deformation of the prosthesis because the prosthesis is defined in a rigid multi-link system. Furthermore, the deformation is limited to rotation around the sagittal plane only. Considering the flexible deformation is challenging compared to a typical rigid multi-link system, however, it is necessary to accurately estimate the ground reaction force when the prosthesis contacts the ground. To account for the viscoelastic properties of the prosthesis as a leaf-spring, methods have been proposed to model the athlete's motion as one-dimensional viscoelasticity using the Spring-loaded inverted pendulum (SLIP) model \cite{murai2018} and as two-dimensional stiffness using the leaf-spring type model \cite{hase2020}. However, these limitations are an obstacle to reproducing the three-dimensional deformation of the prosthesis during activity such as running.\\ \indent
In order to more accurately simulate the physical behavior of a prosthesis during activities such as walking and running, it is necessary to calculate and analyze the three-dimensional flexible deformation of the prosthesis.
The finite element method (FEM) is widely used for the structural analysis of flexible deformation and has been applied to the analysis of soft robots and prosthesis.
Using FEM, the elastic deformation characteristics of a prosthesis can be investigated from the stress distribution when an external force is applied \cite{rahman2014, rigney2017}. However, the purpose of such a study was to investigate the durability of the prosthesis under repeated loading. It doesn't show the dynamic characteristics of the prosthesis under dynamic motion, such as running, where constantly changing loads are applied to the prosthesis in different directions. In addition, the computational cost becomes relatively high as the number of meshes increases, although FEM can handle complex geometries. Therefore, this disadvantage can be an obstacle when analyzing complex movements involving the whole body. \\ \indent
On the other hand, in the field of soft robotics, the Piece-wise Constant Curvature (PCC) model \cite{webster2010} and the Piece-wise Constant Strain (PCS) model \cite{renda2018, renda2018b} have been proposed to calculate the deformation of flexible rod structures.
Compared with FEM, the PCS model can calculate the flexible deformation at a lower computational cost, assuming that it approximates a finite number of segments with the assumption that the strain in each segment is constant.
The PCS model is also a useful method for investigating how elastic structures such as a prosthesis deform and generate forces during motion, as it also shares a common geometric structure with the equations of motion of the rigid multi-link systems such as humanoid and human skeletal models.
The hybrid link system, which integrates a PCS model and a rigid-body multi-link system, has been proposed \cite{ishigaki2021}. In the motion capture calculation of an athlete wearing a prosthesis, the whole body motion including the three-dimensional deformation of the prosthesis during running was reconstructed by inverse kinematics calculation using a skeletal model and a PCS model hybrid link system \cite{kim2022}. 
In addition to this integration, if the dynamic properties of a prosthesis, such as stiffness and viscosity, are known, we can analyze the elastic force stored in the prosthesis during physical activity and its mechanical effects on the body. 
In previous work, we have proposed a method for estimating the stiffness matrix using the PCS model based on the measured deformation of the prosthesis \cite{shimane2022}.
However, this previous work was limited to estimating stiffness from static deformation and force equilibrium only.
Therefore, we need to estimate the viscosity in the dynamics of prosthesis motion.\\ \indent
In the field of robotics, viscoelasticity, consisting of viscosity and stiffness, as well as mass and inertia parameters is identified to accurately calculate the dynamics in terms of manipulator and humanoid control.
The viscoelasticity estimation method has been proposed using inertia parameters obtained from dynamic identification using the equations of motion of the base link \cite{mikami2014}. On the other hand, joint viscoelasticity of human limbs has been estimated using a musculoskeletal model from optimization calculations based on measured motion capture and EMG data \cite{venture2009, fang2017}. However, these studies focused on joint stiffness, defined as a rigid multi-link system, and are inadequate for flexible structures such as a prosthesis. On the other hand, in the numerical calculation of the inverse kinematics of a robot, stable convergence of the calculation with a small number of iterations is achieved by introducing virtual viscoelasticity between the end efector position and the target position, and using elastic energy as the evaluation function and convergence decision \cite{sekiguchi2020}. However, while this study considers virtual springs in its discussion, it does not focus on inverse kinematics calculations for manipulators composed of materials with viscoelastic properties.\\ \indent
In this study, to reproduce the three-dimensional dynamics of the prosthesis, we apply the PCS model to a prosthesis deformation, assuming that it can be divided into a finite number of segments.
In addition, we proposed an energy-based inverse kinematics calculation method that is reasonably consistent with the material properties of the prosthesis by considering a term that minimizes the elastic energy determined from material properties such as Young's modulus and Poisson's ratio. Using this approach, we calculate the reasonable strain of the prosthesis based on optical motion capture data.
Furthermore, we propose a method to estimate the viscosity and stiffness matrix from these calculated strains and ground reaction force data by solving a quadratic programming. We also demonstrate the validity of the results by comparing the forward dynamics simulation result based on the estimated viscoelasticity with the measurement result.

The rest of this paper is organized as follows.
In Section 2, we describe the basic kinematics equations of the PCS model and propose the energy-based inverse kinematics approach.
In Section 3, we describe the dynamics of the PCS model and show the dynamics calculation result compared with the measured ground reaction forces.
In Section 4, we propose the viscoelasticity estimation method and present the experimental result.
Then, in Section 5, we evaluate the estimated viscoelasticity by applying the results to the forward dynamics simulation. 
Finally, we summarize the obtained results and conclude this paper in Section 6.
\section{Inverse kinematics of PCS model}
\subsection{Kinematics of PCS model and its Jacobian matrix \rm{\cite{renda2018}}}
\begin{figure}
   	\centering
	\subfloat[]{\
            \includegraphics[width=0.304\hsize]{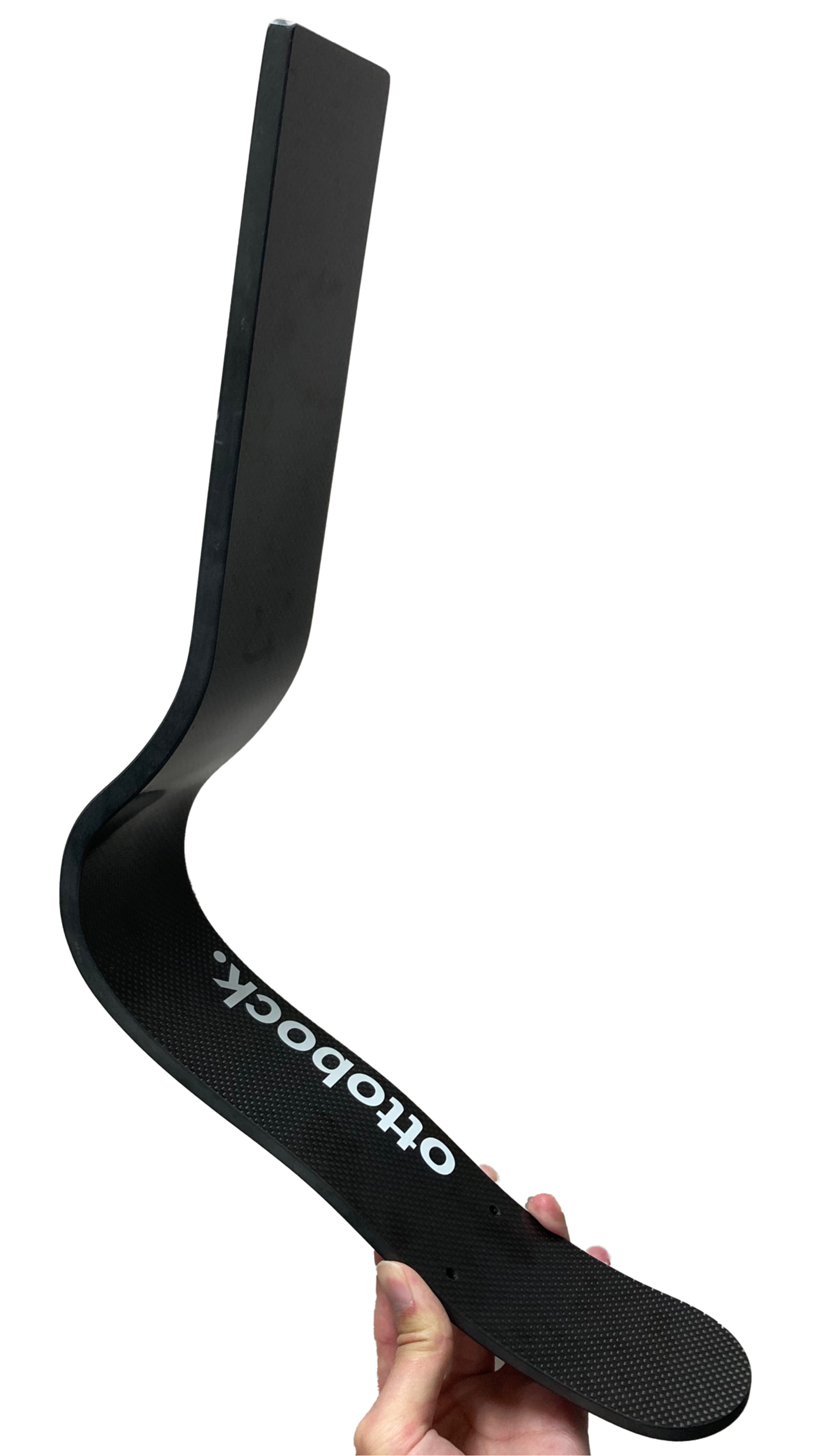}
	}\
	\subfloat[]{\
            \includegraphics[width=0.41\hsize]{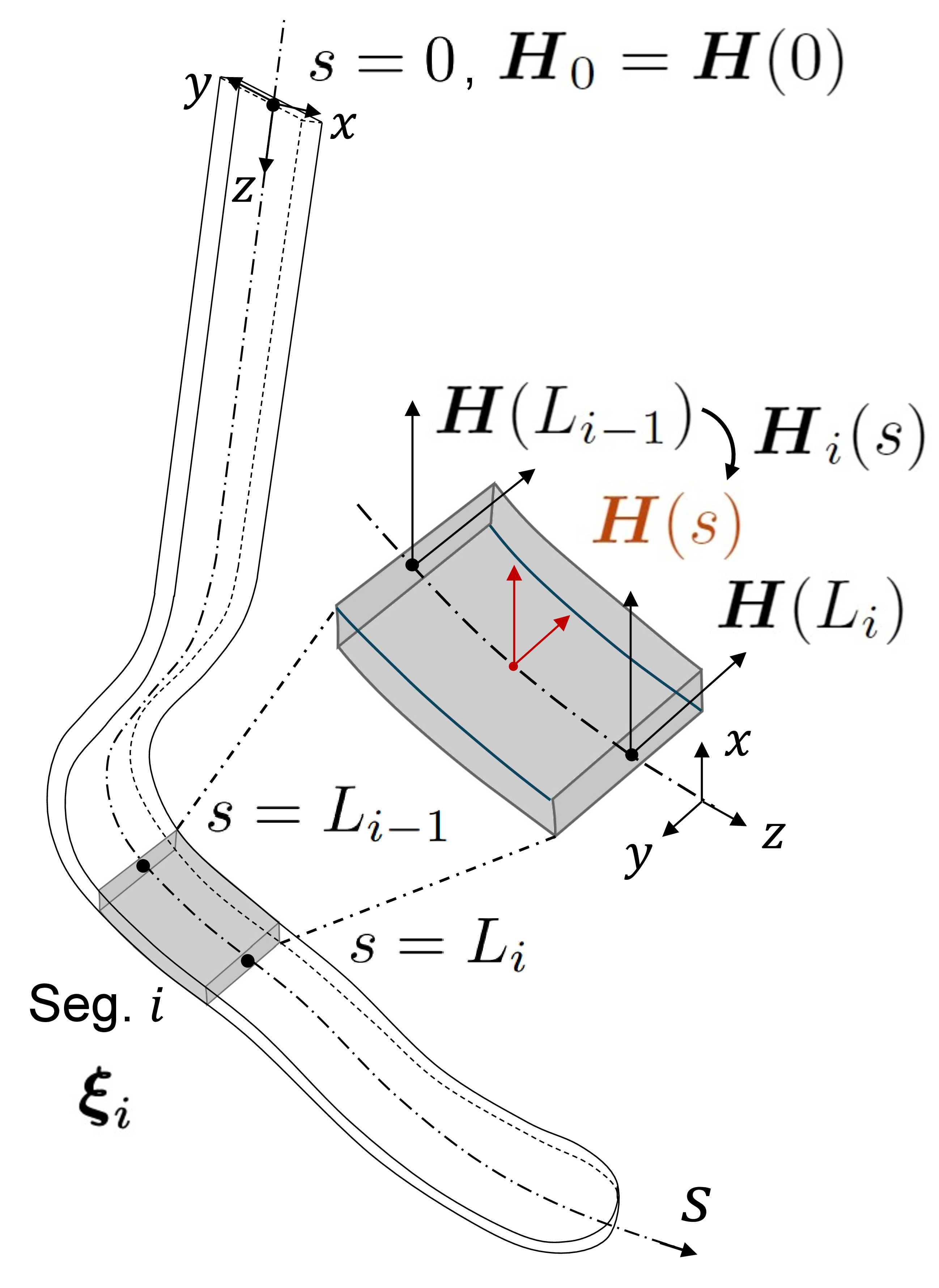}
	}\
         \caption{(a) The sports prosthesis (Sprinter 1E90, Ottobock). (b) Schematic illustration of the PCS model of the sports prosthesis.} 
	\label{fig:prosthesis}
\end{figure}

We use the PCS model to calculate the flexible deformation of the prosthesis shown in Figure \ref{fig:prosthesis} (a). In the PCS model, the deformation of the rod structure is calculated by dividing it into the number of segments based on the continuous Cosserat model \cite{antman2005}, which models soft robots with infinite degrees of freedom. The configuration curve of the PCS model shown in Figure \ref{fig:prosthesis} (b) is defined as follows: 
\begin{align}
 \bm{H}(s)=
 \begin{bmatrix}
  \bm{R}(s) & \bm{p}(s) \\
  \bm{0}^T & 1 
 \end{bmatrix}
 \in \rm{SE}(3),
\end{align}
where $s$ is the material abscissa along the flexible structure such as the prosthesis, $\bm{R}(s)\in\rm{SO(3)}$ and
$\bm{p}(s)\in\mathbb{R}^3$ represent the rotation matrix and the position vector, respectively. 
In the PCS model, the displacement of the configuration curve $\bm{H}(s)$ caused by the flexible deformation is defined as the six-dimensional strain vector $\bm{\xi}(s) = [\bm{k}^T\ \bm{u}^T]^T$ as follows:
\begin{align}
\label{xi}
[\bm{\xi}\times] :=\bm{H}^{-1} \frac{\partial \bm{H}}{\partial s} =
 \begin{bmatrix}
      [\bm{k}\times] & \bm{u} \\
      \bm{0}^T & 0  
 \end{bmatrix} \in \rm{se}(3),
\end{align}
where $\bm{k} \in \mathbb{R}^3$ and $\bm{u} \in \mathbb{R}^3$ represent the angular and linear strains, respectively.
A rod or beam shape such as a prosthesis is divided into a finite number of segments, assuming that the strain $\bm{\xi}$ is constant within each segment.
Defining the $i$-th $(i = 1, \dots, N)$  segment as $L_{i-1} \le s < L_{i}$, the constant strain $\bm{\xi}_i$ of $i$-th segment is defined as follows:
\begin{align}
 \bm{\xi}_i:= \bm{\xi}(s) \ \  (L_{i-1} \le s < L_{i}),
\end{align}
where $N$ is the number of segments, $L_i$ represents the material abscissa at the boundary of the $i$-th segment as shown in Figure \ref{fig:prosthesis} (b). 
$\bm{H}(L_{i})$ and $\bm{H}(L_{i-1})$ have the following relation using the homogeneous transformation matrix $\bm{H}_i(s)$:
\begin{align}
 \bm{H}(s) = \bm{H}(L_{i-1}) \bm{H}_i(s)
 \label{eq:Gs}.
\end{align}

$\bm{H}_i(s)$ can also be derived from (\ref{xi}) using the exponential mapping as follows:
\begin{align}
 \bm{H}_i(s) := \exp\{(s-L_{i-1})[\bm{\xi}_i\times]\}
 \label{eq:Gis}.
\end{align}

\ \ The time variation of the configuration curve $\bm{H}(s)$ is defined by a twist vector $\bm{\eta}(s) = [\bm{\omega}^T\ \bm{v}^T]^T \in \mathbb{R}^6$ as follows:
\begin{align}
 [\bm{\eta}\times] :=\bm{H}^{-1} \frac{\partial \bm{H}}{\partial t}=
 \begin{bmatrix}
  [\bm{\omega}\times] & \bm{v} \\
  \bm{0}^T & 0  
 \end{bmatrix}
 \in \rm{se}(3),
\end{align}
where $\bm{\omega}$ and $\bm{v}$ are the angular and linear velocities, respectively. The velocity twist $\bm{\eta}(s)$ is calculated at the $i$-th segment as follows:
\begin{align}
\label{eq:eta}
 \bm{\eta}(s) &= \mathrm{Ad}^{-1}_{\bm{H}_i(s)} (\bm{\eta}(L_{i-1})+\bm{T}_i(s)\dot{\bm{\xi}}_i),
\end{align}
where $\bm{T}_i(s)$ is the tangent operator of the exponential map and $\mathrm{Ad}_{\bm{H}}\in\mathbb{R}^{6\times6}$ is the adjoint representations of SE(3) as follows:
\begin{align}
    \bm{T}_i(s)\coloneqq \int_{L_{i-1}}^{s} \mathrm{Ad}_{\bm{H}_i(u)}du, \quad 
    \mathrm{Ad}_{\bm{H}}=
    \begin{bmatrix}
        \bm{R} & \bm{O}\\
        [\bm{p}\times]\bm{R} & \bm{R}
    \end{bmatrix}.
\end{align}

The equation (\ref{eq:eta}) can be expressed as follows by a backtracking to $\bm{\eta}(L_{i-1})$:
\begin{align}
\label{eq:eta2}
    \bm{\eta}(s) &= \sum^i_{j=1} \mathrm{Ad}^{-1}_{\bm{H}^{-1}(L_{j-1})\bm{H}(s)}\bm{T}_j(\min(L_j, s)) \dot{\bm{\xi}}_j 
\end{align}

If we assume that the base segment is fixed to the environment, the generalized coordinate of the PCS model with the strain vector of each segment is defined as:
\begin{align}
\label{eq:qs}
 \bm{q}_{\rm{s}} =
\begin{bmatrix}
 \bm{\xi}_1^T & \bm{\xi}_2^T & \cdots & \bm{\xi}_N^T
\end{bmatrix}^T.
\end{align}

In this paper, we consider a floating-base system as a more general model, where the base segment is not fixed to the environment. 
This system is suitable for attaching the prosthesis to a human skeletal model \cite{kim2022}, or for calculating the ground reaction force by the floating-base dynamics that will be represented by (\ref{eq:eom}). In the floating-base system, the generalized coordinate and the generalized velocity can be represented by $\bm{q}=\{\bm{H}_0,\ \bm{q}_{\rm{s}}\}$ and $ \dot{\bm{q}} = [\bm{\eta}_0^T \ \dot{\bm{q}_{\rm{s}}}^T]^T$, where $\bm{H}_0$ and $\bm{\eta}_0$ represent the homogeneous matrix and velocity vector of base segment with respect to the world frame.

Based on the fact that $\bm{\eta}$ is represented in the task space by the generalized velocity $\dot{\bm{q}}_{\rm{s}}$ as (\ref{eq:eta2}) and (\ref{eq:qs}), the velocity of the floating-base is calculated in the same form as the differential kinematics of the traditional rigid-body multi-link system by the Jacobian matrices of the PCS model $\bm{J}_{\rm{s}}(s, \bm{q}_{\rm{s}})=[\bm{J}_1 \  \cdots  \ \bm{J}_i \cdots \  \bm{J}_N]\in\mathbb{R}^{6\times6N}$ \cite{renda2018} and the base segment $\bm{J}_{0}$ as follows:
\begin{align}
\label{eq:jacobian}
 &\bm{\eta}(s) = \bm{J}(\bm{q})\dot{\bm{q}}, \quad \quad
 \bm{J}(\bm{q}) = 
 \begin{bmatrix}
     \bm{J}_{0} & \bm{J}_{\rm{s}}(s, \bm{q}_{\rm{s}})
 \end{bmatrix},\\
 &\bm{J}_i\coloneqq
    \left\{
    \begin{array}{ll}
         \mathrm{Ad}^{-1}_{\bm{H}^{-1}(L_{j-1})\bm{H}(s)}\bm{T}_j(\min(L_j, s)) & (j \le i) \\
         \bm{O} & (j>i). 
    \end{array}\nonumber
    \right.
\end{align}

\subsection{Inverse kinematics minimizing elastic energy}
In general, a method of the inverse kinematics finds generalized coordinates that minimize the error between the desired position and the current position.
In this study, the three-dimensional positions of the markers are provided by optical motion capture measurements, therefore the minimization problem is as follows:
\begin{align}
    \min_{\bm{q}} \ g(\bm{q}) = \sum_{j=1}^{M} \frac{1}{2}||\bm{p}_{\rm{ref}, \it{j}} - \bm{p}_j(\bm{q})||^2_{W_{j}}
    \label{eq:min},
\end{align}
where $||\bm{x}||^2_{W} = \bm{x}^T\bm{W}\bm{x}$ denotes a weighted squared norm of a vector $\bm{x}$. 
$\bm{W}_{j}$ is the weight matrix, $\bm{p}_{\rm{ref}, \it{j}}$ and $\bm{p}_j$ are the reference position such as the measured optical marker position and the current marker position in the PCS model of the $j$-th $(j=1, \dots, M)$ marker, respectively.

The inverse kinematics calculations use the positional information from the optical motion capture markers placed on the prosthesis, but there are modeling errors associated with the position of these attachments. 
In addition, the fact that the PCS model assumes that strain is constant within a segment introduces the modeling error.
The inverse kinematics calculation is sensitive to these modeling errors in order to compensate for the corresponding marker position errors. As a result, a large strain could be calculated even in the direction of higher stiffness.

To overcome these problems, we consider the elastic energy minimization based on material properties as a reasonable inverse kinematics calculation, and add the elastic energy term to (\ref{eq:min}) as follows:
\begin{align}
    \min_{\bm{q}} \ g(\bm{q}) &= \sum_{j=1}^{M} \frac{1}{2}||\bm{p}_{\rm{ref}, \it{j}} - \bm{p}_j(\bm{q})||^2_{W_{j}} + \frac{\alpha}{2}||\varDelta \bm{q}_{\rm{s}}||^2_{\widehat{K}},\nonumber\\
    \label{eq:min2}
    \varDelta \bm{q}_{\rm{s}} &:= \bm{q}_{\rm{s,eq}} - \bm{q}_{\rm{s}},
\end{align}
where $\alpha>0$ is the appropriate weight coefficient for the energy term.
$\bm{q}_{\rm{s,eq}}$ is the value of the generalized coordinate of the PCS model with the unloaded state.
In a highly stiff structure such as a sport prosthesis, the translational strain $\bm{u}$ is assumed to be negligible compared to the angular strain $\bm{k}$. Therefore, we consider only the angular strain with respect to $\bm{q}_{\rm{s}}$ unless otherwise noted in the following.

$\widehat{\bm{K}} = \rm diag\{\bm{K}_{1}, \dots, \bm{K}_{\textit{N}}\}$ is the stiffness matrix, where $\rm diag\{\bm{x}\}$ is the diagonal matrix.
The stiffness matrix of the each segment ${\bm{K}}_{i} \ (i=1, \dots, N)$ is calculated as follows:
\begin{align}
    {\bm{K}}_{i} &= \mbox{diag} \{EJ_x,\  EJ_y,\  \frac{E}{2(1+\nu)}J_z\}l_i,
\end{align}
where $E$ is Young's modulus, $\nu$ is Poisson's ratio and $l_i$ is the length of each segment.
$J_x, J_y, J_z$ are the cross-sectional second moments for bending and torsion, respectively.
In this paper, the value of Young's modulus ($E = 61.6 \rm{GPa}$) and Poisson's ratio ($\nu = 0.36$) are set according to the references \cite{rigney2017} and \cite{hara2019}, respectively.
The cross-sectional second moment is calculated from the measured geometric value.

In this paper we assume that the actual stiffness is not always equal to $\widehat{\bm{K}}$ because the value of $\widehat{\bm{K}}$ is based on the assumption of the PCS model. In the real prosthesis, the strain of each segment is not always constant. In addition, the anisotropy of the Carbon Fibre Reinforced Plastic (CFRP) is not taken into account in the PCS model. Therefore, we used $\widehat{\bm{K}}$ as the weighting matrix in the inverse kinematics and estimated the actual stiffness value from the experiment reported in the next section.

In the numerical inverse kinematics calculation, the desired generalized coordinates $\bm{q}_{k+1}$ can be found from the following iteration at the $k$-th step:
\begin{align}
    \bm{q}_{k+1} = \bm{q}_k -{\nabla^2 g^{-1}(\bm{q}_k)}\nabla g^T(\bm{q}_k).
    \label{eq:ik}
\end{align}

In preparation for computing the gradient vector $\nabla g$, the objective function $g(\bm{q})$ of (\ref{eq:min2}) can be simplified by defining the vector $\bm{e}_k$ as follows:
\begin{align}
    g(\bm{q}_k) = \frac{1}{2} ||\bm{e}_k||^{2}_W, \quad
    \bm{e}_k = 
        \begin{bmatrix}
            \bm{p}_{\rm{ref}, \it{1}} - \bm{p}_1(\bm{q}_k) \\
            \vdots\\
            \bm{p}_{\rm{ref}, \it{M}} - \bm{p}_M(\bm{q}_k) \\
            \varDelta \bm{q}_{\rm{s}, \it{k}}
        \end{bmatrix},
\end{align}
where $\bm{W} = \mbox{diag}\{\bm{W}_{1}, \dots, \bm{W}_{M}, \alpha\widehat{\bm{K}} \}$. Therefore, the gradient vector can be calculated as follows:
\begin{align}
    &\nabla g(\bm{q}_k)  = \left.\frac{\partial g(\bm{q})}{\partial \bm{q}}\right|_{q=q_k} = -\bm{e}_k^T\bm{W}\widehat{\bm{J}}(\bm{q}_k), \nonumber\\
    &\widehat{\bm{J}}(\bm{q}_k) =
    \begin{bmatrix}
         \widehat{\bm{J}}_1^T(\bm{q}_k) &\dots& \widehat{\bm{J}}_M^T(\bm{q}_k) & [\bm{0} & \bm{E}]^T
    \end{bmatrix}^T,
\end{align}
where $\widehat{\bm{J}}_j (j = 1, \dots, M)$ is the Jacobian matrix of $j$-th marker position described in the world coordinate using the Jacobian matrix $\bm{J}(\bm{q})$ at $\bm{H}(s)$ in (\ref{eq:jacobian}) as follows:
\begin{align}
    &\widehat{\bm{J}}(\bm{q}_k) = 
    \begin{bmatrix}
        -[\widehat{\bm{p}}_j \times] & \bm{E}
    \end{bmatrix}\bm{\mathcal{R}}(s)\bm{J}(\bm{q}_k),\\
    &\bm{\mathcal{R}}(s) =
    \begin{bmatrix}
        \bm{R}(s) & 0 \\ 0 & \bm{R}(s)
    \end{bmatrix},
\end{align}
where $\widehat{\bm{p}}_j$ is the relative marker position to $\bm{H}(s)$.

In the Levenberg–Marquardt method \cite{levenberg1944, marquardt1963}, which is a fast and robust algorithm, the Hessian matrix $\nabla^2 g$ can be approximated by the damping factor $\bm{W}_n$ as follows:
\begin{align}
    \nabla^2 g(\bm{q}_k)  = \left.\frac{\partial^2 g(\bm{q})}{\partial^2 \bm{q}}\right|_{q=q_k} \simeq \widehat{\bm{J}}^T(\bm{q}_k)\bm{W}\widehat{\bm{J}}(\bm{q}_k) + \bm{W}_n.
\end{align}

\begin{figure}[t]
	\centering
	\subfloat[Jig for measurement]{
            \includegraphics[width=0.7\hsize]{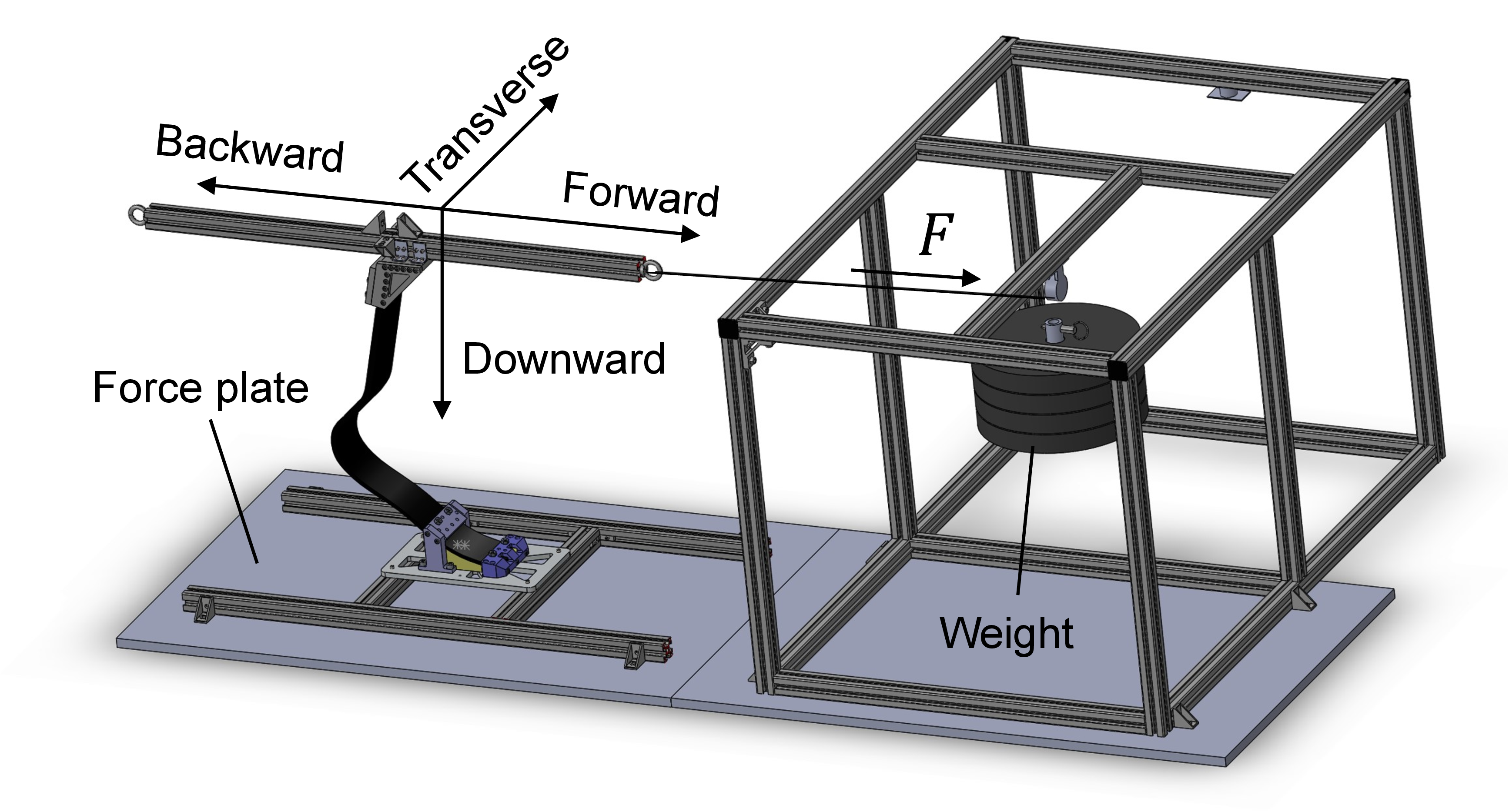}}
        \\
	\subfloat[Marker array \hspace{15mm} (c) Experimental procedure]{
		  \includegraphics[width=0.7\hsize]{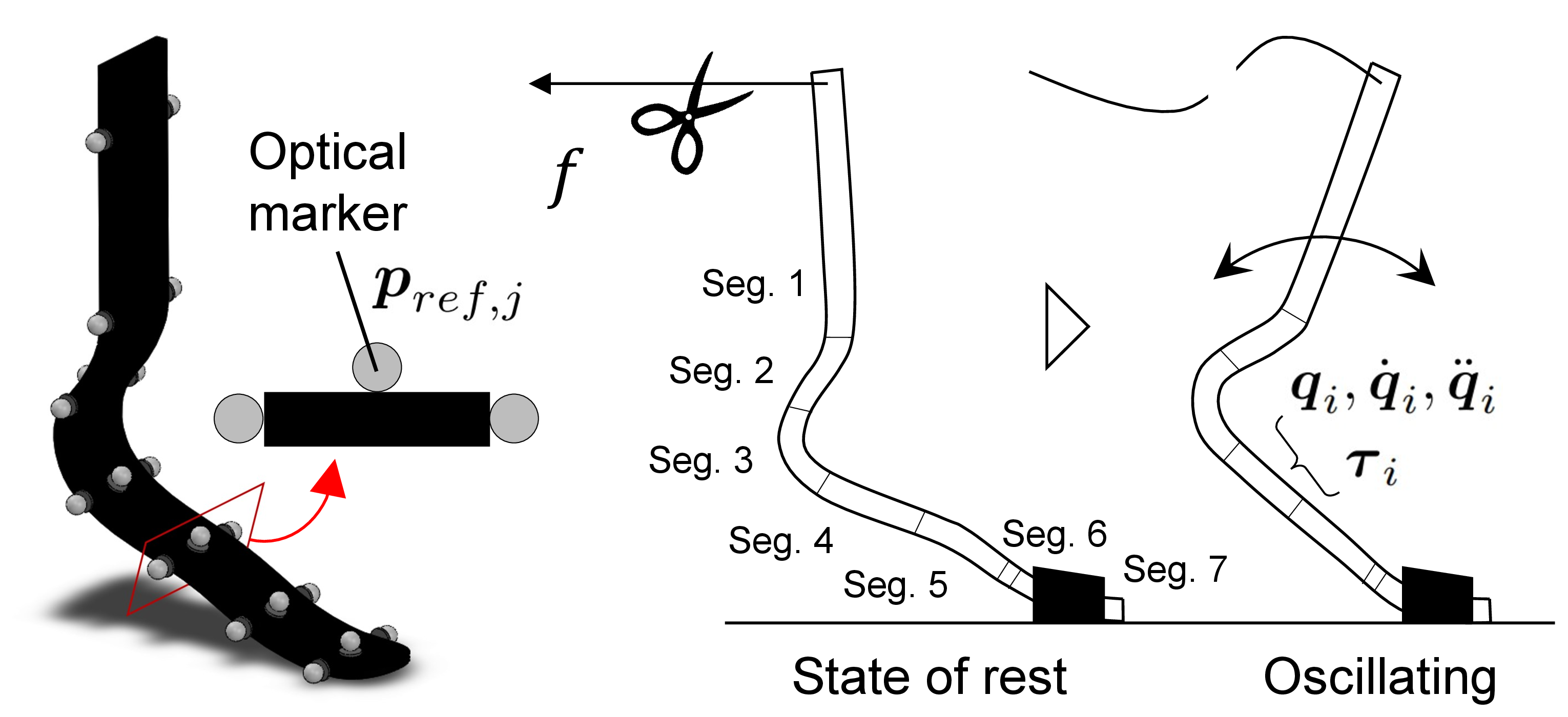}}
         \caption{(a) Measuring jig used in the experiment. The prosthesis was loaded in four different directions to investigate different deformations. (b) Marker array attached to the prosthesis. Twenty-two retro-reflective optical markers were attached. (c) Experimental procedure.} 
	\label{fig:IVA}
\end{figure}

\subsection{Application to motion capture measurement}
To estimate the viscoelasticity of the prosthesis, we measured static and dynamic deformations of the prosthesis with varying loads $\bm{f} \in \mathbb{R}^{6}$ applied to the prosthesis using jig made for the experiment and an optical motion capture system.
The jig was designed in a similar way to the method used in \cite{shimane2022} shown in Figure \ref{fig:IVA} (a). In this experiment, we use the 1E90 Sprinter (Ottobock) made of CFRP.
For the measurement, we used fifteen optical motion capture cameras (Eagle-4, Raptor-4, Motion analysis) and the force plate (Kistler). The sampling rates for each system are 200 Hz and 1000 Hz, respectively. Figure \ref{fig:IVA} (b) shows the arrangement of the twenty-one retroreflective optical markers placed on the prosthesis in the experiment to be evenly distributed in each segment. Three markers were placed on each cross section perpendicular to the material abscissa along the prosthesis. 
Using these equipment, we investigate the displacement of the marker position as deformation after momentary release of the weights as shown in Figure \ref{fig:IVA} (c). 
The prosthesis was loaded in different directions: forward direction, backward direction, transverse direction to investigate various deformations.
We determined these values of loads referring to the joint torque acting on the prosthesis during running \cite{murai2018}. 

\begin{figure}[t]
    \begin{minipage}[t]{1.0\hsize}
    \centering
    \includegraphics[width=0.7\hsize]{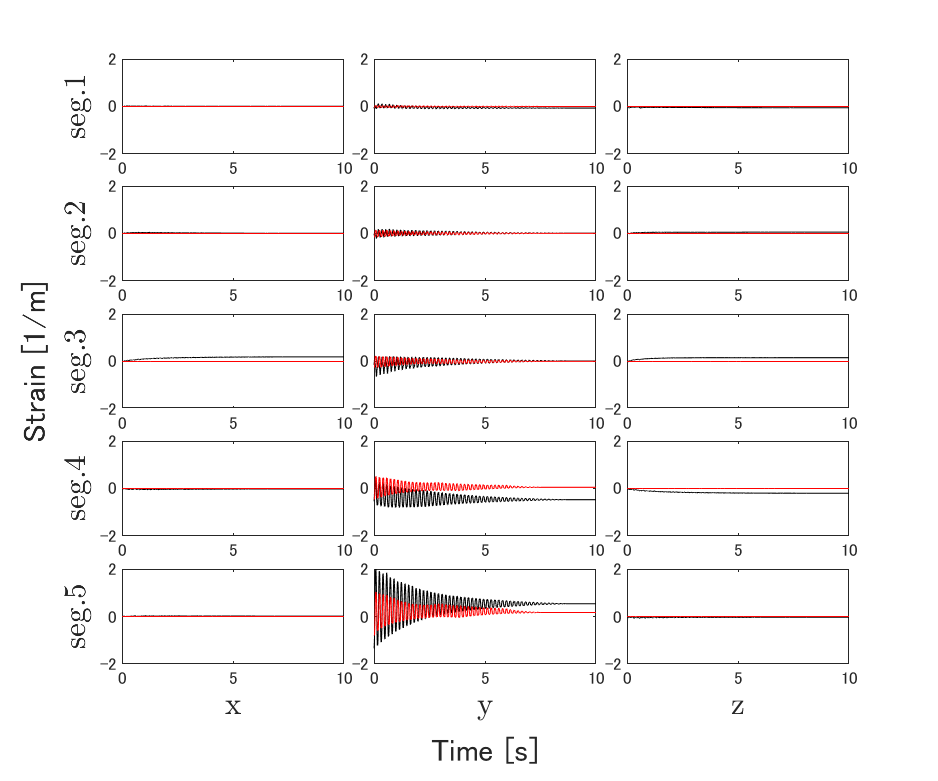} 
    \caption{Results of the strain calculated by the inverse kinematics: The red line is the result of the proposed approach and the black line is the result of the conventional inverse kinematics calculation.} 
    \label{fig:ik_graph}
    \end{minipage}
    \begin{minipage}[t]{1.0\hsize}
    \centering
    \subfloat[Conventional inverse kinematics]{
    \includegraphics[width=0.45\hsize]{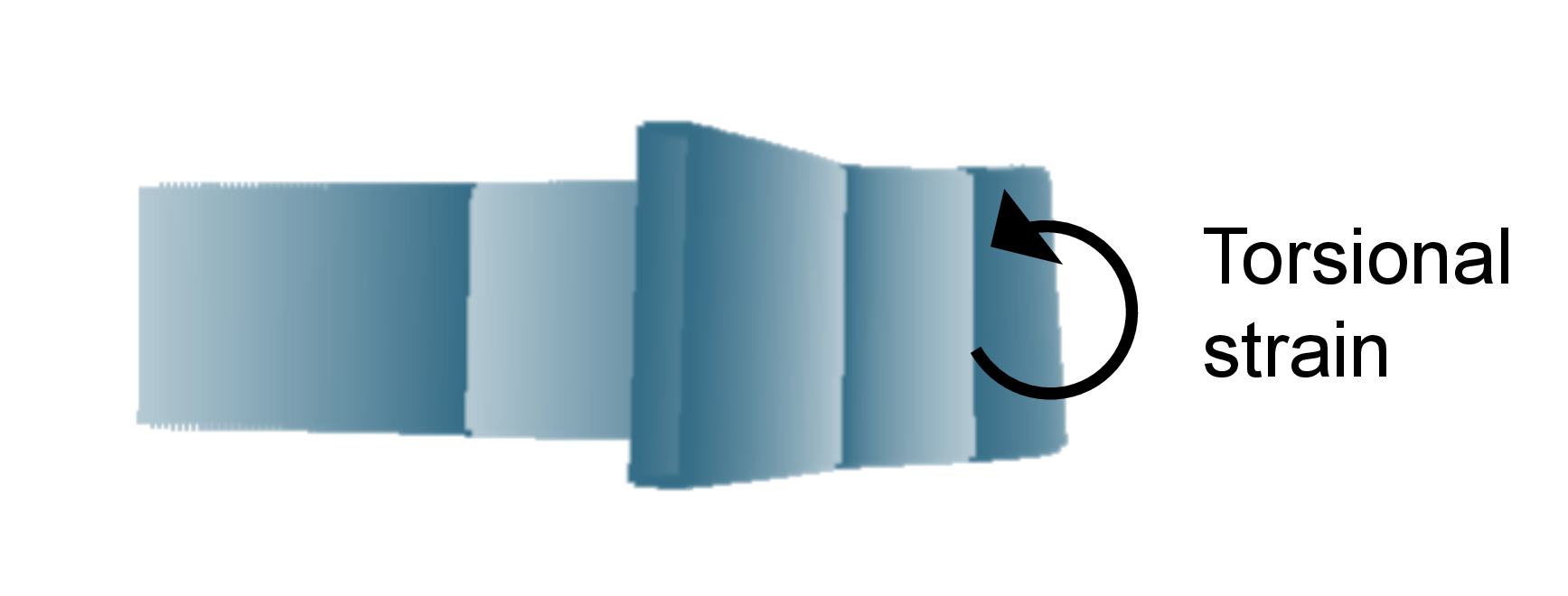}}
    \subfloat[Proposed inverse kinematics]{
    \includegraphics[width=0.32\hsize]{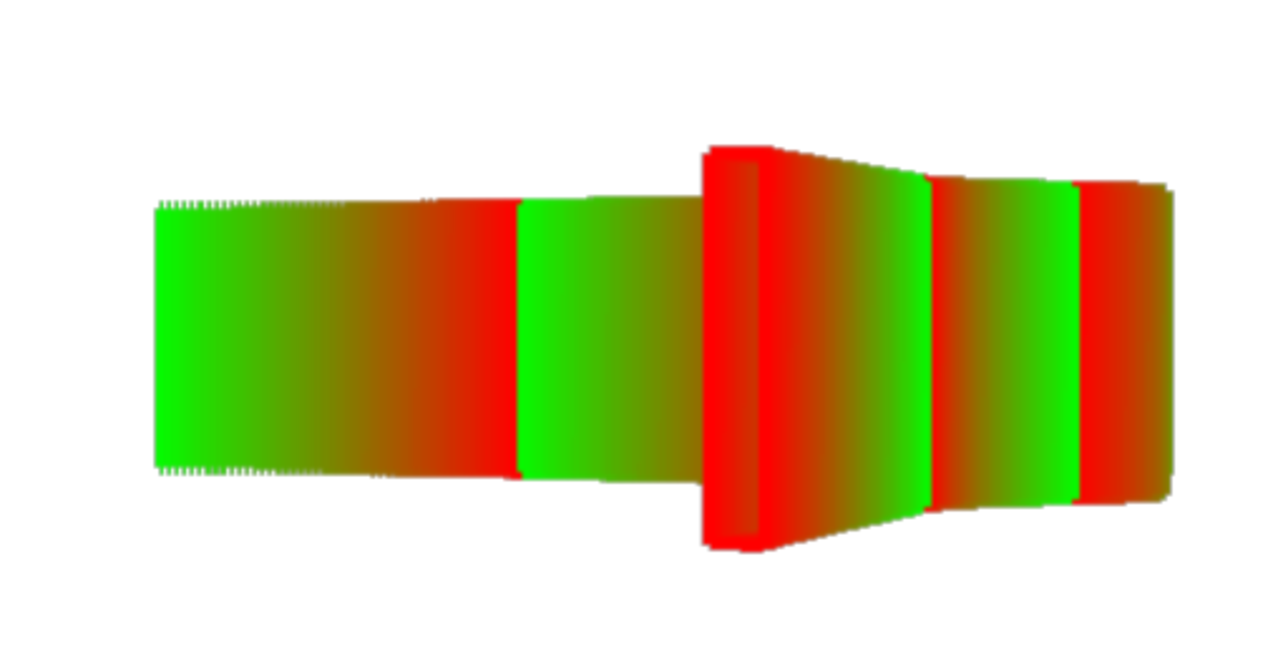}}
    \caption{The PCS model of the prosthesis deformed according to the result of the strain: In the first segment, the conventional inverse kinematics has an inconsistent strain, while the result of the proposed inverse kinematics is more reasonable.} 
    \label{fig:straindraw}
    \end{minipage}
\end{figure}

Based on these marker position data, we obtain the generalized coordinate $\bm{q}$ from the inverse kinematics calculation. 
In this calculation, we use the PCS model of the 1E90 sprinter which is reconstructed in three dimensions with seven segments by selecting locations with large curvature changes as the segment boundaries according to the method used in \cite{shimane2022}.
However, the sixth and seventh segments are fixed to the ground by the jig as shown in Figure \ref{fig:IVA} (a), therefore we assume that these segments are rigid bodies with no degrees of freedom.
The PCS model of the prosthesis is shown in the next section in Figure \ref{fig:straindraw}.

\subsection{Inverse kinematics result of strain}
Figure \ref{fig:ik_graph} shows the inverse kinematics calculation results for the motion data of the backward horizontal direction with 15kg weight according to the order of the segments as shown in Figure \ref{fig:IVA} (a).
The red line is the result of the proposed approach and the blue line is from a conventional inverse kinematics calculation that does not include the energy minimization term. 

In the conventional calculation method, the strain value is not zero when the damping oscillation converges, and the oscillation is shifted towards the horizontal axis. 
Furthermore, focusing on the z-direction in Figure \ref{fig:ik_graph}, the strain occurred in the stiffer torsional direction as shown in Figure \ref{fig:straindraw} (a), even though the prosthesis is applied with forces acting only in the direction of the main bending and moving in forward/backward direction. These inconsistent results, which seem to differ from the real physical behavior, are improved by using the proposed approach.
\section{Dynamics of PCS model of prosthesis}
\subsection{Dynamics of the floating base PCS model}
In preparation for the estimation of viscoelasticity and the calculation of dynamics, we explain the dynamics of the PCS model of the prosthesis.
The equation of motion of the floating base segment can be represented as follows:
\begin{align}
    \begin{bmatrix}
      \bm{M}_0 & \bm{M}_{0\rm{s}}\\
      \bm{M}_{0\rm{s}}^T & \bm{M}_{\rm{s}}
    \end{bmatrix}
    \begin{bmatrix}
      \dot{\bm{\eta}}_0 \\ \ddot{\bm{q}}_{\rm{s}}
    \end{bmatrix}+
    \begin{bmatrix}
      \bm{b}_0 \\ \bm{b}_{\rm{s}}
    \end{bmatrix}=
        \begin{bmatrix}
      \bm{0} \\ \bm{\tau}_{\rm{s}}
    \end{bmatrix}+
    \begin{bmatrix}
        \bm{J}^T_{0} \\
        \bm{J}^T_{\rm{s}}
    \end{bmatrix}\bm{f}
    \label{eq:eom},
\end{align}
where $\bm{M}_0, \bm{M}_{0\rm{s}}, \bm{M}_{0\rm{s}}$ and $\bm{M}_{\rm{s}}$ are inertial matrices, $\bm{b}_0$ and $\bm{b}_{\rm{s}}$ are bias vectors including the Coriolis force and gravity. In this study, the weight of the prosthesis (0.59kg) and jig (forward/backward direction: 1.0kg, transverse direction:1.82kg) were measured and used to calculate the dynamics.
$\bm{f} \in \mathbb{R}^{6}$ is an external force. In the situation of the motion capture measurement, $\bm{f}$ represents the ground reaction force. 
$\bm{J}_{0}, \bm{J}_{\rm{s}}$ is the contact Jacobian assuming that the ground reaction force acts on the prosthesis on the material abscissa.
The subscripts $0$ and $\rm{s}$ denote the quantity of the base segment and the PCS model segments, respectively. 
$\bm{\tau}_{\rm{s}} \in \mathbb{R}^{6N}$ is the generalized force, which is the viscoelastic internal force represented as:
\begin{align}
    \bm{\tau}_{\rm{s}} = \bm{K}\Delta\bm{q}_{\rm{s}} - \bm{D}\dot{\bm{q}}_{\rm{s}},
    \label{eq:tau_s}
\end{align}
where $\bm{K} \in \mathbb{R}^{3N \times 3N}, \bm{D} \in \mathbb{R}^{3N \times 3N}$ are stiffness and viscosity matrices, respectively. 

\subsection{Ground reaction force estimation}
Given the generalized acceleration $\ddot{\bm{q}}=[\dot{\bm{\eta}}_0^T, \ddot{\bm{q}}_{\rm{s}}^T]$ from the result of the inverse kinematics calculation described in the previous section, we can estimate the ground reaction force using the upper half of (\ref{eq:eom}).
Since the matrix $\bm{J}_0$ is regular, we can calculate the ground reaction force $\widehat{\bm{f}}$ as follows:
\begin{align}
    \widehat{\bm{f}} = \bm{J}_0^{-T} (\bm{M}_{0}^T\dot{\bm{\eta}}_0 + \bm{M}_{0s}\ddot{\bm{q}}_{\rm{s}} - \bm{b}_0).
    \label{eq:force_h}
\end{align}

The value in the brackets on the right-hand side of (\ref{eq:force_h}) can be calculated by inverse dynamics for given $\bm{q}, \dot{\bm{q}}$ and $\ddot{\bm{q}}$. 

\subsection{Result of ground reaction force estimation}
\begin{figure*}[t]
  \centering
    \includegraphics[width=1.0\hsize]{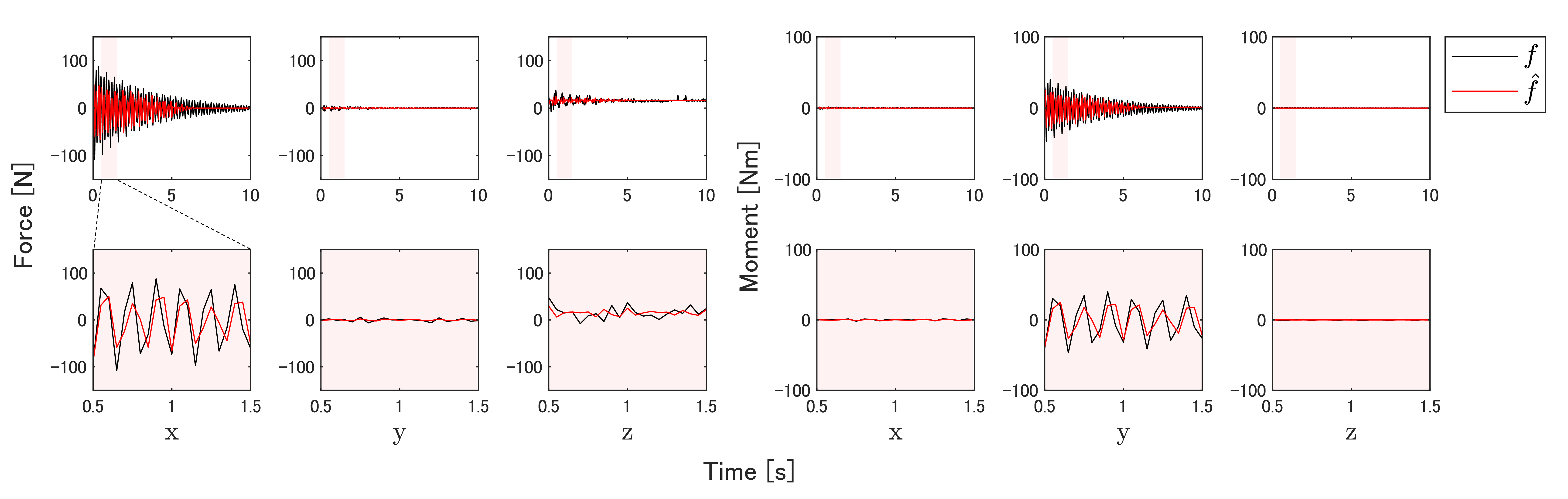}
  \vspace*{-2mm}
  \caption{Comparison of ground reaction force values in the backward horizontal direction: The red line represents the estimated value $\widehat{\bm{f}}$ based on the equation of motion, and the black line represents the measured value $\bm{f}$. The shaded area is magnified view of the one second period.}
  \label{fig:id_force}
\end{figure*}

Figure \ref{fig:id_force} shows the comparison of the estimated ground reaction force $\widehat{\bm{f}}$ with the measured value $\bm{f}=[\bm{F}, \bm{n}]$ in one of the measured data which is shown in Section 2.4, where $\bm{F} = [F_x, F_y, F_z]$ and $\bm{n} = [n_x, n_y, n_z]^T$ are the force and moment, respectively. 
The black and red lines represent ground reaction forces $\bm{f}$ and $\widehat{\bm{f}}$ respectively.
The force $F_x$ and moment $n_y$ are larger than the other elements due to the mainly bending direction in this measurement. 
As a result, the measured and estimated values are qualitatively similar.
Finally, the Root Mean Square Errors (RMSE) of the measured and estimated ground reaction forces for $\bm{F}$ and $\bm{n}$ are calculated respectively as follows:
\begin{align}
    \sqrt{\frac{1}{m} \sum^{m}_{j=1} ||\bm{y} - \widehat{\bm{y}}||^2},
    \label{eq:rmse}
\end{align}
where the force $\bm{F}$ or moment $\bm{n}$ is selected for $\bm{y}$ to calculate the RMSE. The RMSEs in foward horizontal direction are 9.2 \% and 7.3 \% of the data range for force and moment, respectively.
The results demonstrate the validity of the kinematic and kinetic calculations based on the prosthetic PCS model, and the feasibility of reproducing the dynamic behavior of the prosthesis based on the base segment in the upper half of (\ref{eq:eom}).

\section{Viscoelasticity estimation}
In the PCS model \cite{renda2018}, the stiffness and viscosity matrix is defined by the material properties, such as Young's modulus and Poisson's ratio based on the assumptions of the PCS model. 
However, the material such as CFPR of leaf-spring type prosthesis has anisotropic properties, and the whole shape of the prosthesis, which is composed of curved surfaces, can affect the viscoelasticity.
Therefore, the viscoelasticity calculated by such an approach does not sufficiently correspond to the actual characteristics of the prosthesis.
For this reason, we estimate the viscoelasticity based on the measurement data and the result of the kinematics and dynamics calculation in the following.

\subsection{Estimation of stiffness and viscosity matrix}
The stiffness and viscosity matrix are estimated substituting (\ref{eq:tau_s}) into the lower part of (\ref{eq:eom}):
\begin{align}
    \bm{\Gamma} =
     \bm{M}_{0\rm{s}}^T\dot{\bm{\eta}}_0 + \bm{M}_{{\rm{s}}}\ddot{\bm{q}}_{\rm{s}} - \bm{b}_{\rm{s}} - \bm{J}_{\rm{s}}^T\bm{f} = \bm{K}\Delta \bm{q}_{\rm{s}} - \bm{D}  \dot{\bm{q}}_{\rm{s}}.
\end{align}

In this equation, the left-handside can be calculated by the inverse dynamics.
The stiffness $\bm{K}$ and the viscosity matrix $\bm{D}$ are estimated by the following optimization.
\begin{align}
\label{eq:obj}
 &\min_{\bm{X}} \ \frac{1}{2} \sum_{l=1}^{n}\sum_{j=1}^{m} ||\bm{\Gamma}_{lj} 
 - \bm{K} \varDelta \bm{q}_{\rm{s}, \it{lj}} + \bm{D} \dot{\bm{q}}_{\rm{s}, \it{lj}}||^2_{W_{lj}},
 \nonumber
 \\
 &\mbox{s.t.} \ \bm{X} \succeq 0,
\end{align}
where $\bm{q}_{\rm{s}, \it{lj}}$ and $\bm{\Gamma}_{lj}$ are the values at $j$-th ($j=1,2,\dots, m$) frame of $l$-th ($l=1,2,\dots, n$) measurement data.
$\bm{W}_{lj}$ is the weight matrix. 

$\bm{X}$ is determined by whether the stiffness matrix $\bm{K}$ or the viscosity matrix $\bm{D}$ is to be estimated.
First, the stiffness matrix is estimated based on the static deformations when a constant load is applied to the prosthesis ($\bm{X} = \bm{K}$), which can be consider as $\dot{\bm{q}}_s=0, \ddot{\bm{q}}_s=0$.
Then, the viscosity matrix is then estimated from the dynamic deformations in oscillation due to the restoring forces mentioned in section 2.4, given the known stiffness.

In the previous study \cite{shimane2022}, we defined and estimated the stiffness matrix as the block diagonal matrix considering that the deformation of anisotropic materials such as CFRP may have force and direction interference. As a result, we found that the diagonal component of the elastic matrix is dominant. Therefore, we define the stiffness matrix $\bm{K} \succeq 0$ and the viscosity matrix $\bm{D} \succeq 0$ as the following semi-positive definite diagonal matrix based on the measured motion data of the prosthesis:
\begin{align}
    \label{diagK}
     \bm{K} = \mbox{diag}\{\bm{K}_i \}, \quad \bm{D} = \mbox{diag}\{\bm{D}_i \},
\end{align}
where $\bm{K}_i, \bm{D}_i$ are the stiffness and viscosity matrices of $i$-th segment. 

Finally, we solve the minimization problem (\ref{eq:obj}) as a quadratic programming using MATLAB (The MathWorks).

\begin{figure*}[t]
    \begin{minipage}[t]{1.0\hsize}
        \centering
        \includegraphics[width=1.0\hsize]{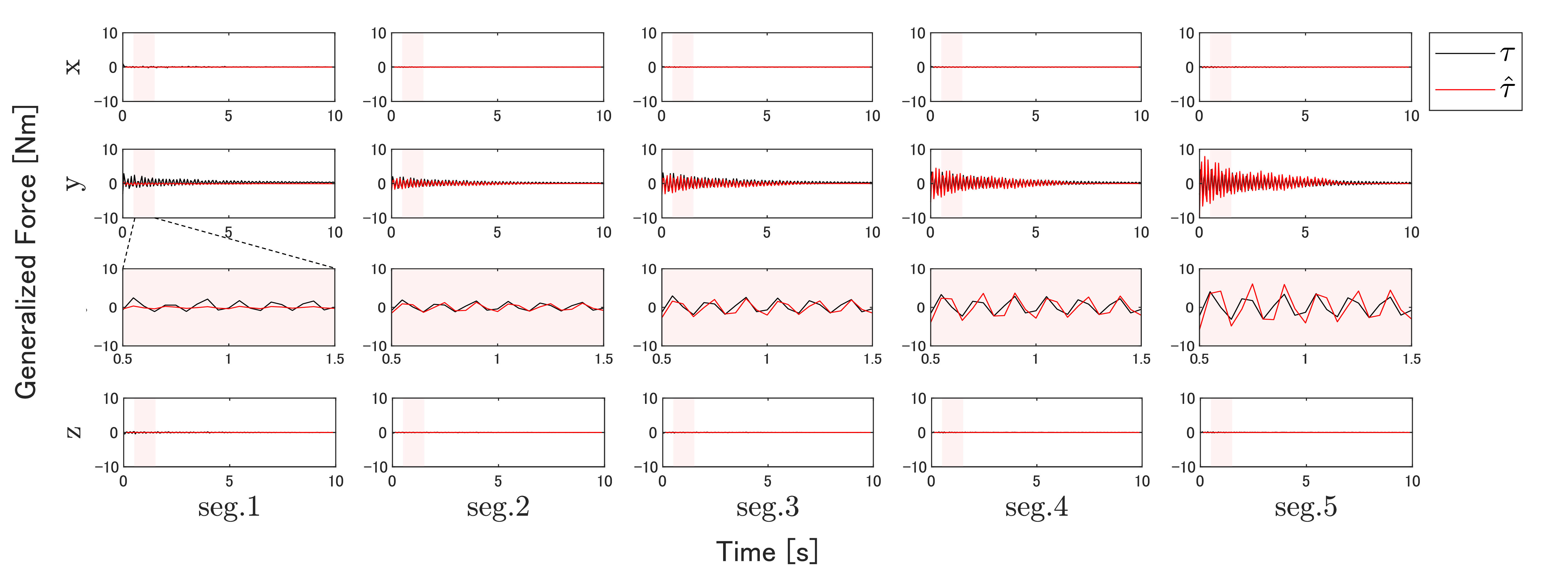}
        \vspace*{-2mm}
        \caption{Comparison of measured and estimated generalized force of the prosthesis PCS model: The red line represents the estimated value using the estimated stiffness and viscosity matrix, and the black line represents the measured value calculated from the measured force. The shaded area is magnified view of the one second period.}
        \label{fig:id_genforce}
    \end{minipage}\\
    \begin{minipage}[t]{1.0\hsize}
        \centering
        \subfloat[Stiffness matrix]{
                \includegraphics[width=0.7\hsize]{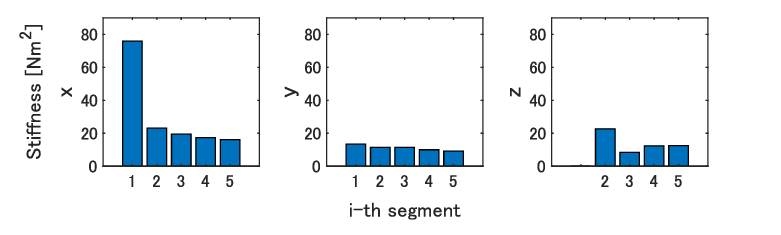}}
            \\
        \subfloat[Viscosity matrix]{
              \includegraphics[width=0.7\hsize]{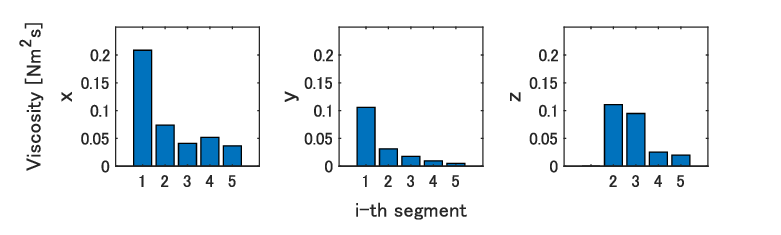}}	
             \caption{The estimated result of (a) the stiffness matrix $\bm{K}$ and (b) the viscosity matrix $\bm{D}$ sorted with respect to the x-y-z strains.} 
        \label{fig:stivis_graph}
    \end{minipage}
\end{figure*}

\subsection{Result of viscoelasticity estimation}
Figure \ref{fig:stivis_graph} shows the element of the stiffness matrix $\bm{K}\in \mathbb{R}^{15 \times 15}$ and the viscosity matrix $\bm{D}\in \mathbb{R}^{15 \times 15}$ sorted with respect to the x-y-z strains.
Figure \ref{fig:stivis_graph} (a) shows the stiffness matrix, where the stiffness for the y-axis is less than that for the x- and z-axes.
This indicates that the stiffness for the y-axis is relatively soft and easy to bend, which is the main bending direction of the prosthesis. 
However, this estimation did not include any measurement data that would cause strain in the torsional direction of the first segment in the z-direction, where the stiffness is considered to be stiffer than the other segments, therefore it was excluded from the estimation.

Figure \ref{fig:stivis_graph} (b) shows the viscosity matrix, where the viscosity for the y-axis is similarly smaller than that for the x- and z-axes in each segment.
This shows that the strain in the transverse and torsional directions of the prosthesis causes a large damping force.
In fact, the transverse oscillation shown in the next section converges faster than the forward oscillation in Figure \ref{fig:ik_graph} of Section 2.4.

Using these estimated stiffness and viscosity matrices, the estimated generalized forces are compared with the value calculated from the measured value $\bm{f}$ as the following equations of motion:
\begin{align}
    \bm{J}_{\rm{s}}^T {\bm{f}} = \bm{M}_{0\rm{s}}^T\ddot{\bm{q}}_0 + \bm{M}_{\rm{s}}\ddot{\bm{q}}_{\rm{s}} - \bm{b}_{\rm{s}}  - \bm{K}\Delta \bm{q}_{\rm{s}} + \bm{D}\dot{\bm{q}}_{\rm{s}} +\bm{\tau}_{0}.
    \label{eq:tau}
\end{align}

Figure \ref{fig:id_genforce} shows the result and comparison of the generalized force.
The red line represents the estimated value of right-hand side of (\ref{eq:tau}), and the black line represents the value of left-hand side. 
The consistency of the curves between the estimated and measured values can be qualitatively confirmed.
\section{Forward dynamics simulation}
To evaluate the estimated viscoelasticity and the modeling of the prosthesis, we verify whether the actual behavior can be reproduced by the forward dynamics simulation.
Once the viscoelasticity and the prosthesis strain estimated in the previous section are known, we can calculate the restoring and damping forces, which allows us to calculate the forward dynamics. 
Here, we consider the simulation of the same motion as in the measurement experiment used to estimate the viscoelasticity.
The measured data not used to estimate the viscoelasticity in the previous section are used for verification.  
A lateral load of 35kg is applied to the prosthesis and released momentarily.

\begin{figure}[H]
    \begin{minipage}[t]{1.0\hsize}
      \centering
    \includegraphics[width=1.0\hsize]{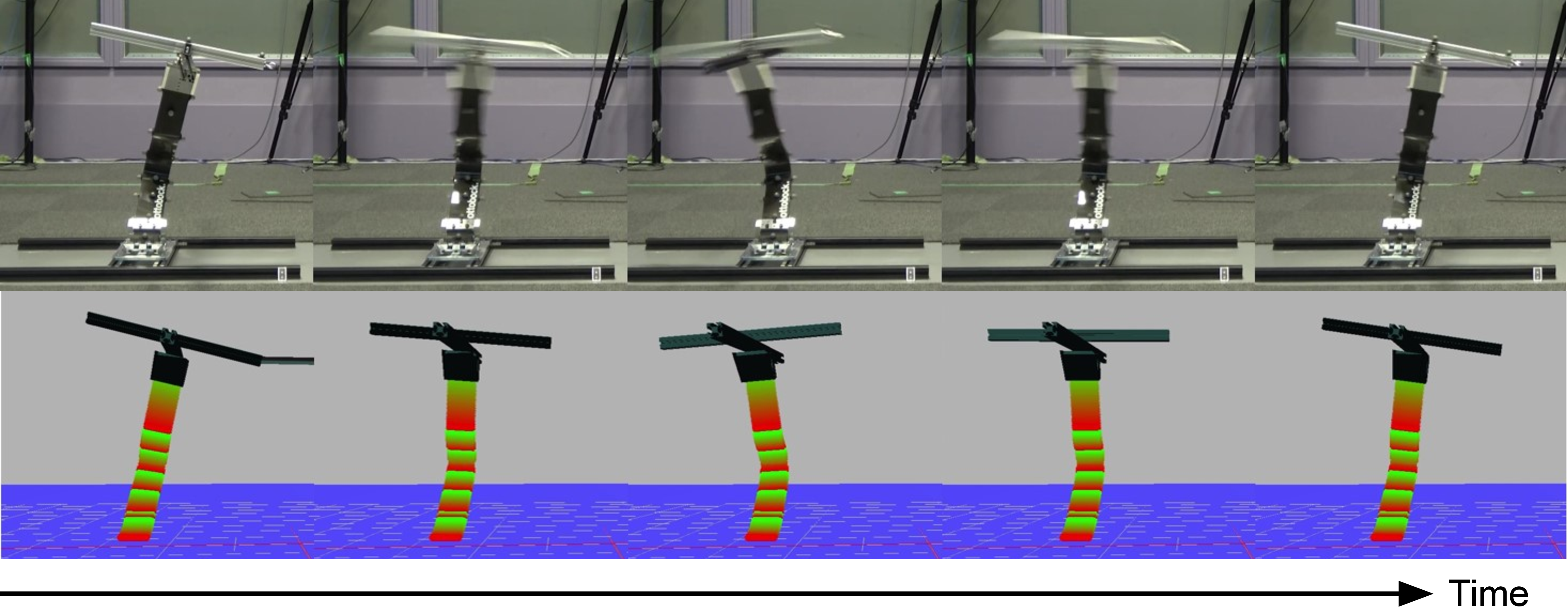}
      \vspace*{-2mm}
      \caption{Results of experimental measurement and forward dynamics calculation in transverse horizontal direction: Oscillatory motion of the prosthesis after release of the load applied to the upper part by the jig can be confirmed. The prosthesis is pulled by a wire and bent to the right, statically (leftmost figure). It then moves gradually to the left due to the elasticity of the prosthesis.}
      \label{fig:fd_photo}
    \end{minipage}
    \begin{minipage}[t]{1.0\hsize}
    \centering
    \includegraphics[width=0.8\hsize]{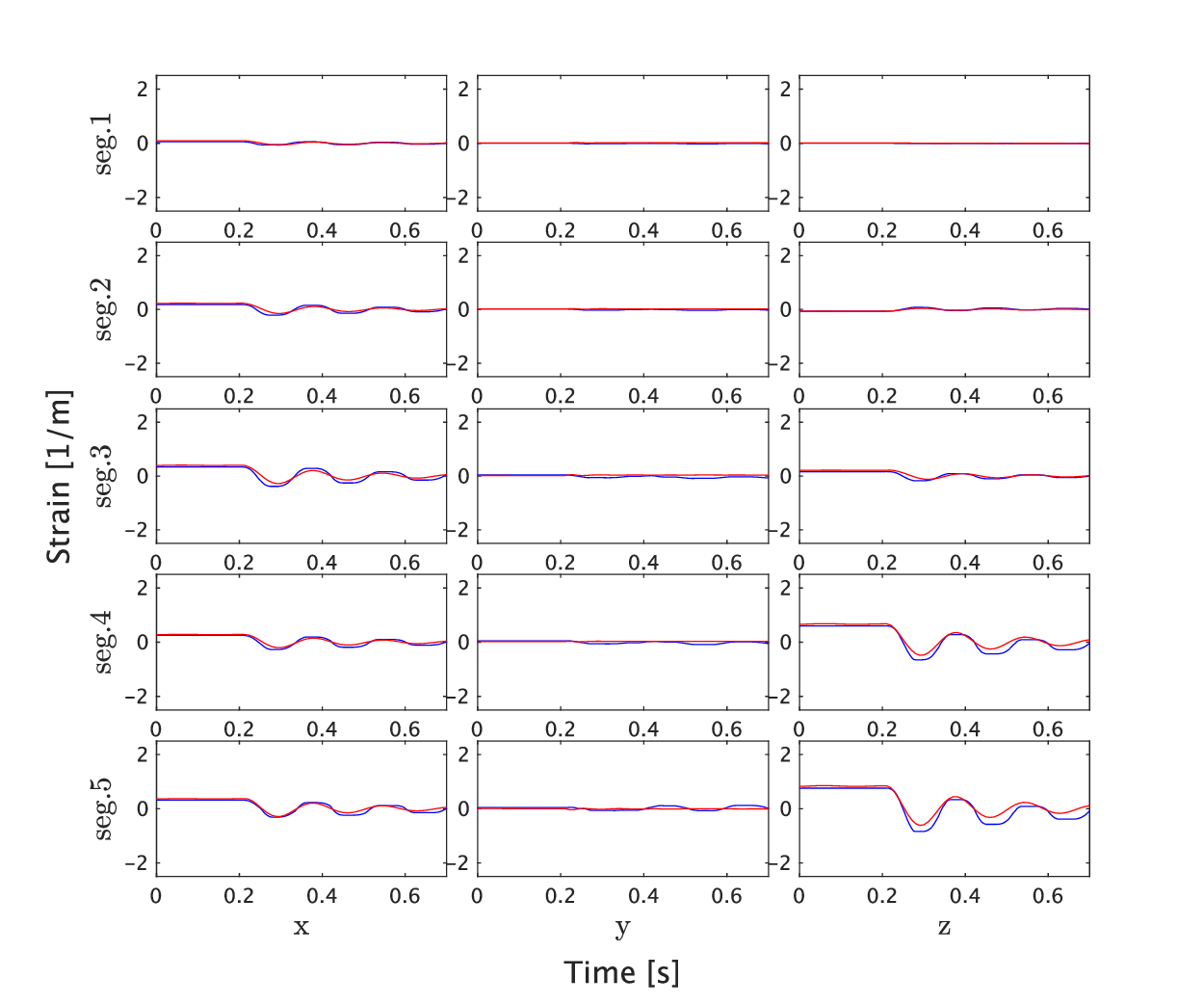} 
    \caption{Results of the strain by the forward dynamics: The red line shows the result of the forward dynamics simulation, which reproduces the oscillation phenomena after applying 35kg to the side of the prosthesis. The blue line shows the inverse kinematics calculation result based on the measured data.} 
    \label{fig:fd_graph}
    \end{minipage}
\end{figure}
Figure \ref{fig:fd_photo} shows the results of the forward dynamics calculation in the transverse horizontal direction compared with the experimental measurement.
We can see the oscillatory motion of the prosthesis after releasing the load applied by the jig on the upper part. 

Figure \ref{fig:fd_graph} shows the strain calculated based on the measured data by inverse kinematics calculation and the results of the forward kinematics calculation.
The strains obtained from the inverse kinematics and forward dynamics calculations are almost in perfect agreement in the static section during the period up to 0.2s when the prosthesis is under external force.
This is because the stiffness properties, which are the relationship between the strain and the restoring force of each segment, are correctly modeled for the external force acting on the prosthesis, reproducing the behavior of the prosthesis in a static situation.

In the dynamic section after 0.2s, when no force is applied to the prosthesis, there are some errors in the strain due to viscoelasticity estimation and modeling errors, but the waveforms are similar in terms of cycle and magnitude.
Finally, the RMSE between the both was about up to 12$\%$ in the z-axis direction for the sixth segment.
This indicates that the viscoelasticity is correctly estimated and therefore the mechanical oscillation of the prosthesis is reproduced.
\section{Conclusion}
In this paper, we present the Piece-wise Constant Strain (PCS) model of the sports prosthesis in order to reproduce the three-dimensional dynamics of the prosthesis. 
We proposed a reasonable inverse kinematics calculation method that is consistent with the physical properties of the prosthesis.
In addition, we proposed a method to estimate the stiffness and viscosity of the prosthesis based on the measured motion capture data. 
Furthermore, we verify the proposed method and the viscoelasticity estimation result by forward dynamics simulation.
The results of this study are summarized as follows:
\begin{enumerate}
\item 
In order to accurately calculate the flexible deformation of a prosthesis in three dimensions, we proposed a reasonable inverse kinematics calculation method using the PCS model. The conventional inverse kinematics calculation methods, which are based on geometric optimization of marker positions, cause unreasonable strain in the calculation of flexible deformation of elastic materials.
In contrast, the proposed method improves these problems by taking into account elastic energy minimization based on the material properties of the prosthesis and is able to calculate reasonable deformations.
\item
We proposed the viscoelasticity estimation method to reproduce the dynamic behavior of a prosthesis. The viscoelasticity matrix was estimated using quadratic programming based on several measured motion capture data of the prosthesis.
The estimation result indicated the three-dimensional viscoelastic characteristic of a leaf-spring type prosthesis made of materials such as the CFRP, including the overall shape of the prosthesis.
\item 
In order to evaluate the proposed method and the estimation result of viscoelasticity, 
we simulated the motion of the prosthesis by forward dynamics calculations based on the estimation result of viscoelasticity under the same conditions as the experimental measurements.
By comparison with the strain calculated directly from the measured motion data, we showed that our method reproduced the actual three-dimensional dynamic behavior of the prosthesis with a maximum RMSE of 12$\%$.
In conclusion, our approach allows us to simulate the motion of the athlete wearing the prosthesis under different conditions on a computer.
\end{enumerate}\par 

\section*{Funding}
This work was supported by the JSPS KAKENHI under Grant number 21H01282; and JST SPRING under Grant number JPMJSP2108.

\bibliographystyle{tfnlm}
\bibliography{ref}

\end{document}